\documentclass[10pt,twocolumn,letterpaper]{article}

\usepackage{cvpr}
\usepackage{times}
\usepackage{epsfig}
\usepackage{graphicx}
\usepackage{amsmath}
\usepackage{amssymb}
\usepackage{tabularx}
\usepackage{bbding}
\usepackage{array}
\usepackage{amssymb}
\usepackage[normalem]{ulem}

\cvprfinalcopy 

\begin{document}

\title{Enabling Pedestrian Safety using Computer Vision Techniques: A Case Study of the 2018 Uber Inc. Self-driving Car Crash}

\author{Puneet Kohli\\
Texas A\&M University\\
College Station, TX USA\\
{\tt\small puneetkohli@tamu.edu}
\and
Anjali Chadha\\
Texas A\&M University\\
College Station, TX USA\\
{\tt\small anjali\_chadha@tamu.edu}
}

\maketitle

\begin{abstract}
Human lives are important. The decision to allow self-driving vehicles operate on our roads carries great weight. This has been a hot topic of debate between policy-makers, technologists and public safety institutions. The recent Uber Inc. self-driving car crash, resulting in the death of a pedestrian, has strengthened the argument that autonomous vehicle technology is still not ready for deployment on public roads. In this work, we analyze the Uber car crash and shed light on the question, "Could the Uber Car Crash have been avoided?". We apply state-of-the-art Computer Vision models to this highly practical scenario. More generally, our experimental results are an evaluation of various image enhancement and object recognition techniques for enabling pedestrian safety in low-lighting conditions using the Uber crash as a case study.
\end{abstract}


\section{Introduction}

   Nearly 1.3 million people die in road crashes each year, averaging to 3,287 deaths a day. Another 20-50 million people are injured or disabled. These statistics are alarming, and if no action is taken, road accidents would be the fifth leading cause of death by 2030 \cite{asirt}. Human error, such as distraction, fatigue, speeding, and misjudgment -- have been shown to be responsible for 90\% of road accidents \cite{blincoe2015economic}. One of the promises of Autonomous Vehicle (AV) technology is to significantly mitigate this impending public health crisis \cite{anderson2014autonomous, FAGNANT2015167}. Autonomous vehicles can not get distracted, drunk, or tired. An additional boost in performance for AVs comes from the advances in artificial intelligence, sensor fusion, and computer vision techniques that essentially \textit{self-drive} the vehicle. Despite this, we have seen through various incidents that self-driving technology is not even near perfect. A recent study shows that self-driving cars would have to be driven hundreds or millions of miles to accurately demonstrate their safety \cite{KALRA2016182}. Many companies and research institutions alike have started deploying self-driving cars and autonomous vehicles on the roads for testing in an arms race to secure a foothold on this upcoming market.
   
   The Uber Inc. Self-Driving Car Crash \footnote{VentureBeat - Uber self-driving car crash in Tempe, Arizona. Youtube Video: https://youtu.be/XtTB8hTgHbM}{https://youtu.be/XtTB8hTgHbM} which occurred in Tempe, Arizona (USA) was one of the most recent accidents involving self-driving cars which unfortunately resulted in the demise of a pedestrian. The crash involved a Volvo XC90 sport utility vehicle in fully autonomous mode hitting a pedestrian as she was crossing the road at night in a generally low visibility environment. It is shocking and still unknown as to why Uber's systems were unable to detect the pedestrian before the crash happened. Although reports claim that the car's system was unable to detect the pedestrian due to the lighting conditions, a study by Intel shows that their proprietary MobileEye system was able to detect the pedestrian one second before the crash happened \cite{mobileye}.
   
   On similar lines, we evaluate the performance of various state-of-the-art Object Recognition frameworks, both neural network based and traditional machine learning based, to detect the pedestrian in the post-crash dashboard camera (dash-cam) footage released by Uber. We also perform a variety of image enhancement techniques in a best-effort approach to detect the pedestrian sooner than with just the raw footage. In Section 3, we first go through the image enhancement techniques we applied on the video, followed by the object recognition techniques we used. Section 4 and 5 discuss our experimental setup and findings in-depth. Section 6 shows our vision for the future prospects of this work.

\begin{figure*}[t]
\centering
\includegraphics[width=1.0\textwidth, ]{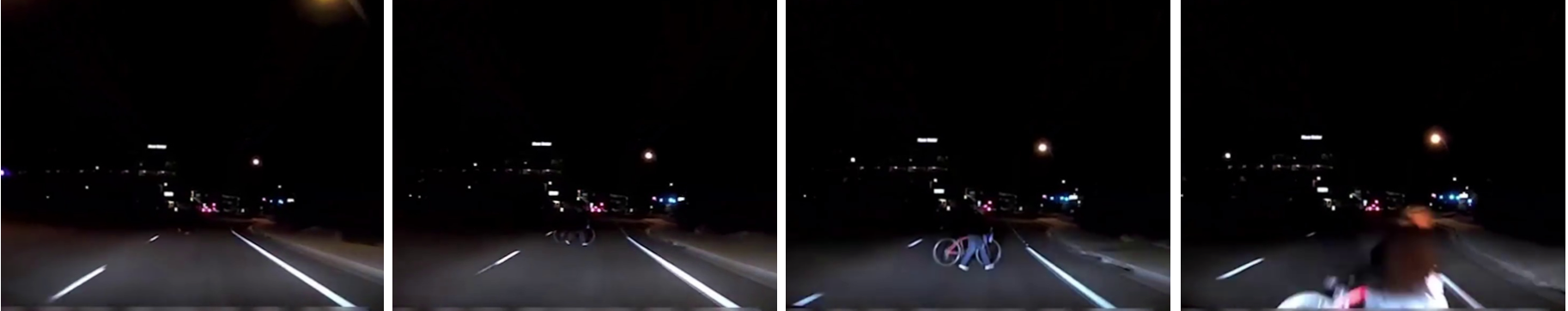}
\label{fig:testsfigure7}
\caption{Frames 60, 70, 80, and 90 of the dash-cam footage released by Uber Inc.}
\end{figure*}
\section{Related Work} \label{lit}

Over the last decade, multiple researchers have studied the problem of pedestrian detection \cite{LRPD100, LR2, LR3, LR4, LR5haar}. However, the current systems aren't sophisticated enough to be deployed in the self-driving cars without human behind the wheel. This is a challenging task because of the difference in humans' postures, appearances.  The problem discussed here is two-fold --- pedestrian detection and image enhancement. Therefore, we will discuss the prior work related to the two topics in separate sections.
\subsection{Pedestrian Detection}
Previous works on this topic can be broadly classified into two categories --- classical models using handcrafted features and those using neural networks for pedestrian detection.
 
    \subsubsection{Classical Methods} 
    
    Earlier work relied on sliding window techniques for object detection using proposal generation or Histogram of Orientated Gradients  (HOG) \cite{HOG_SVM} for feature extraction followed by use of classification methods  such as SVM \cite{LSsvm} and Adaptive Boosting \cite{LRAdaptiveBoosting} for pedestrian detection. 
    
    Before the recent advancements in neural networks, decision forests (boosted) used to give best performance for pedestrian detection. SquaresChnFtrs \cite{10yrsPD}, InformedHaar \cite{LR5haar} and, SpatialPooling \cite{LRpaisitkriangkrai2014strengthening} are some of the top performing boosted decision trees for integral channel features architecture \cite{LR3}. Multiple works also investigated the performance of features extracted from motion to detect human \cite{PDPeriodicity, PDDetectingActivities, PDRealTimeMotionDetection}. While others experimented with the models combining motion and intensity features \cite{PDMotionAppearance}. These hand-crafted features have been widely used for object detection. 
    
    Use of edge detection techniques for pedestrian detection have shown good results as well \cite{Edge5, Edge6, Edge7}. An edge is a boundary between background and an object. There are different categories of edge detectors - gradient \cite{EdgeGradient1, EdgeGradient2, Edge4}, zero crossing \cite{EdgeZeroCrossing}, Laplacian of Gaussian \cite{Edge4, EdgeZeroCrossing}, Gaussian Edge \cite{Edge4}, and Colored Edge detector \cite{EdgeColor}. In our study, we use the classical Canny Edge Detection approach \cite{canny1987computational}.
    
    Partial occlusion of the pedestrian images is another hurdle for human detection. In fact, around 70\% of the pedestrians captured in street scenes are partially occluded in at least one frame of the video \cite{pedestrian_detection_benchmark}. Previous works have tried to tackle this problem by specifically training detectors for different types of occlusions \cite{LRocclusion6, LRocclusion7}, or by modeling part visibility as latent variables \cite{LRocclusion8, LRocclusion9, LRocclusion10, LRPD11}. Mathias et al. trained classifiers for specific types of occlusion, for instance bottom-up or right-left occlusions \cite{LRocclusion7}. Other works divided the pedestrian into multiple parts and inferred their visibility with latent variables \cite{LRPD11, LRocclusion9}. A different class of part-based pedestrian detectors assume that the head of the pedestrian is always visible \cite{wu2005detection}. 

    \subsubsection{Neural Network Methods}
    
    With the recent advances in neural networks, Convolutional Neural Networks have been successfully applied for the object detection \cite{LRfasterRcnn, yoloV2, yolov1, LRFRCNN,  SVM_CNN_traffic_light, R-Cnn}. Recent works also focus on using convnets to achieve build better pedestrian detection models \cite{LRfasterRcnnPD, LRScaleAwareFRCNN, LRStrongParts, LR4, tian2015pedestrian}. Zhang \textit{et al.} \cite{LRfasterRcnnPD} explored the use of Region Proposal Networks (RPN) \cite{LRfasterRcnn} for generating pedestrian candidate boxes followed by a cascaded boosted forest \cite{LRBoosting} for classifying whether a candidate box is a pedestrian. Tian \textit{et al.} \cite{tian2015pedestrian} joins the the task of pedestrian detection with semantic tasks such as pedestrian attributes (example, carrying a backpack). Convnet \cite{LR4} applied convolutional sparse coding to unsupervised pre-trained CNNs for pedestrian detection.
    
    Researchers have tried to combine the classic methods of occlusion handling in pedestrian images \cite{LRocclusion7} with the novel deep learning \cite{LRStrongParts} architectures to achieve better results. In contrast to \cite{LRocclusion7} where the body parts were pre-defined, \cite{LRStrongParts} determines these parts automatically from data which may vary with the datasets and the scenarios. This step is followed by training an ensemble of individual part detectors for pedestrian detection. One advantage of this model over Part-based RCNN model \cite{LRzhang2014part} is that it doesn't require part annotations in training.
    
    \subsubsection{Datasets} 
    
    Multiple pedestrian detection datasets have been made publicly available over the years. ETH \cite{ess2008mobile}, INRIA \cite{HOG_SVM}, Caltech-USA \cite{pedestrian_detection_benchmark, pedestrian_detection_evaluation} , KITTI \cite{kitti}, and TUD-Brussels \cite{wojek2009multi} are a few of the more popular of them. Different datasets vary in the environmental settings (urban, city, mountains), capturing angle (rear-view; stereo rig mounted on  a stroller), diversity of the data (mutliple cities/countries, crowded areas), and the inclusion of partially/occluded pedestrians of varied sizes and postures. We evaluate our model using the Caltech Pedestrian Dataset as it consists of footage recorded from a vehicle driving through an urban scenario with regular traffic conditions which is fitting for our use case.
    
\begin{figure}[t]
\centering
\includegraphics[width=1\linewidth, ]{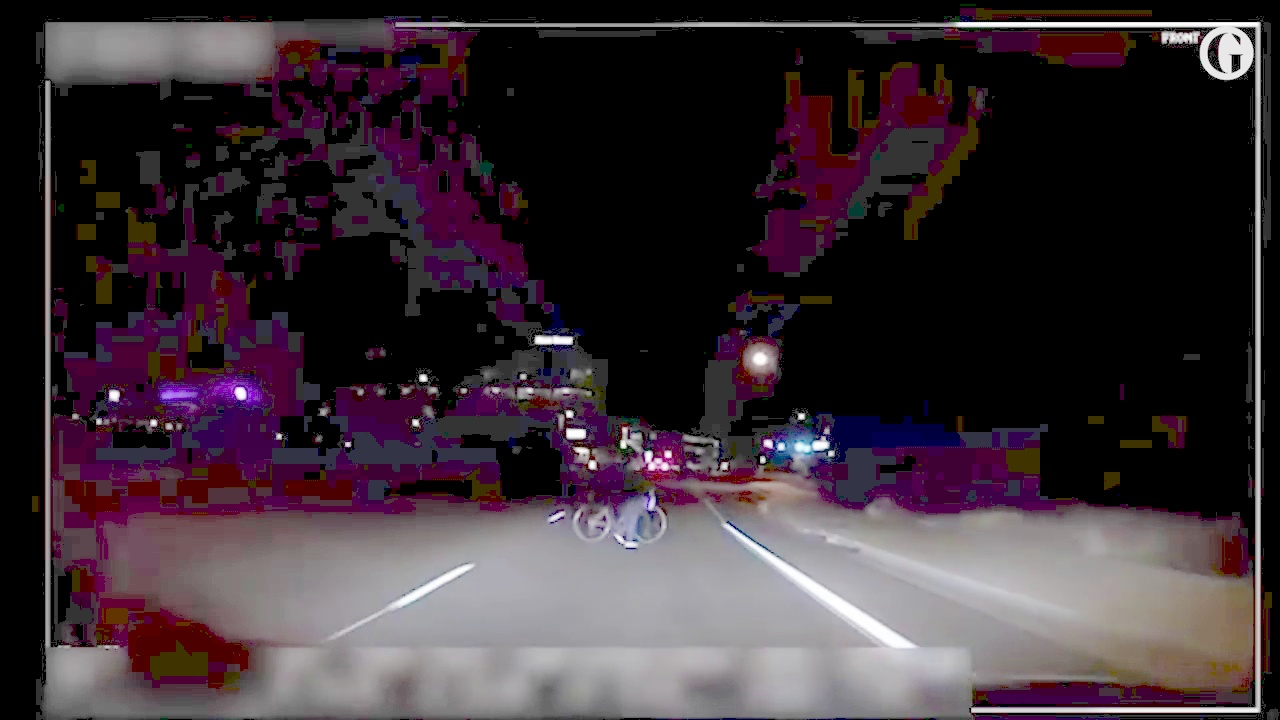}
\label{fig:testsfigure6}
\caption{(Frame 75 after applying Gamma correction ($\gamma = 3.5$)}
\end{figure}

\subsection{Image Enhancement}
One of the major challenges for Computer Vision research and applications is the low quality of input images. Many researchers have worked in the area of Image Enhancement in order to tackle this problem. Our work primarily focuses on pedestrian detection in low-light scenarios. Existing literature for low-light image enhancement can be broadly classified into two groups: histogram based and Retinex based methods.

Histogram equalization \cite{pizer1987adaptive} is one of the most intuitive and common ways to fix the bad lightening in an image. Contrast-limiting Adaptive Histogram Equalization (CLAHE) \cite{HE4clahe} limits the HE result to a certain area. There are other works which explores non-linear functions like gamma function to enhance image contrast. Image denoising tasks have been explored using K-SVD, BM3D and non-linear filters. Retinex methods are based on Retinex Theory introduced by Land \cite{land1977retinex}  to explain the color perception of human vision system. Single Scale Retinex (SSR) is based on center/surround retinex. There are Multi-scale Retinex models \cite{jobson1997multiscale} as well which considers the weighted sum of several different SSR outputs.

There are deep learning models -- LLNet \cite{LLNet}, HDRNet \cite{hdrnet}, MSRNet \cite{MSRNet}, which enhances the image using neural networks. For instance, Fu et al. \cite{fu2017removing} tried to remove rain from single images via deep detailed network. Cai \textit{et al.} \cite{cai2016dehazenet} introduced a trainable end to end system called DehazeNet which takes a hazy image as an input and provides a transmission map which outputs a haze-free image.

\begin{figure}[t]
\centering
\includegraphics[width=1\linewidth, ]{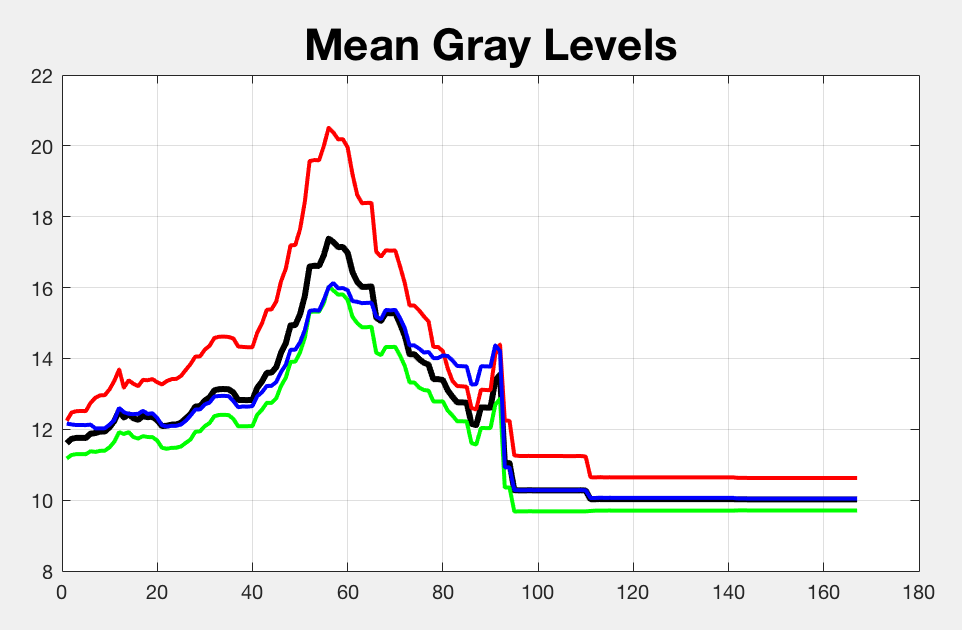}
\label{fig:testsfigure1}
\caption{Grey levels for the RGB channels (depicted in red, green, and blue) for the Uber crash video. The black line represents the mean grey level.}
\end{figure}

\section{Methodology}

Our methodology is divided into two parts. First, the Image Processing and Enhancement Techniques, and second, the Object Detection techniques. We ran our best Object Detection model on both the raw video feed, as well as on videos enhanced via our processing algorithms.

\subsection{Image Processing \& Enhancement}

   \subsubsection{Motion Detection}
   
   One of the first ideas we experimented is detecting motion in the video as in \cite{BackgroungSVM4}. We performed background subtraction in consecutive frames of video. There are multiple methods to perform background subtraction as discussed in \cite{BackgroundMM1, BackgroundAGMM2, BackgroundADE3}. We used a Gaussian Mixture based algorithm for background and foreground segmentation as discussed in \cite{BackgroundAGMM2} which selects the appropriate number of Gaussian distribution for each pixel. This step helps us to segment the background from the foreground. These algorithms are sophisticated enough to distinguish between actual motion and other shadowing or lighting changes in the video. 

    \subsubsection{Gamma Correction}
    
    The notion of gamma stems from the fact that our eye reacts to light in a non linear way. In other words, human eye is more sensitive to small variation in light in dark settings than in a bright one. Gamma Correction is a non-linear operation used to correct image's luminance. Each pixel in a image has brightness level called luminance. This value varies between 0 and 1 where 0 mean complete darkness and 1 is brightest. Gamma correction is also known as \textit{Power Law Transform}.
    \[ V_{out} = V_{out} ^ {1/\gamma}\] 
    $\gamma < $1 shifts the image towards the darker end of the spectrum while $\gamma > $1 makes the image appear lighter. \\
    This technique is often used in night vision systems -- \cite{gamma1, gamma2} apply gamma correction for autonomous/ driver assistance vehicles. To illuminate our video, we tried various $\gamma$ values. For $\gamma=$3.5, we can see the one of the frames of the video in Figure 2.
\begin{figure}[t]
\centering
\includegraphics[width=1\linewidth,]{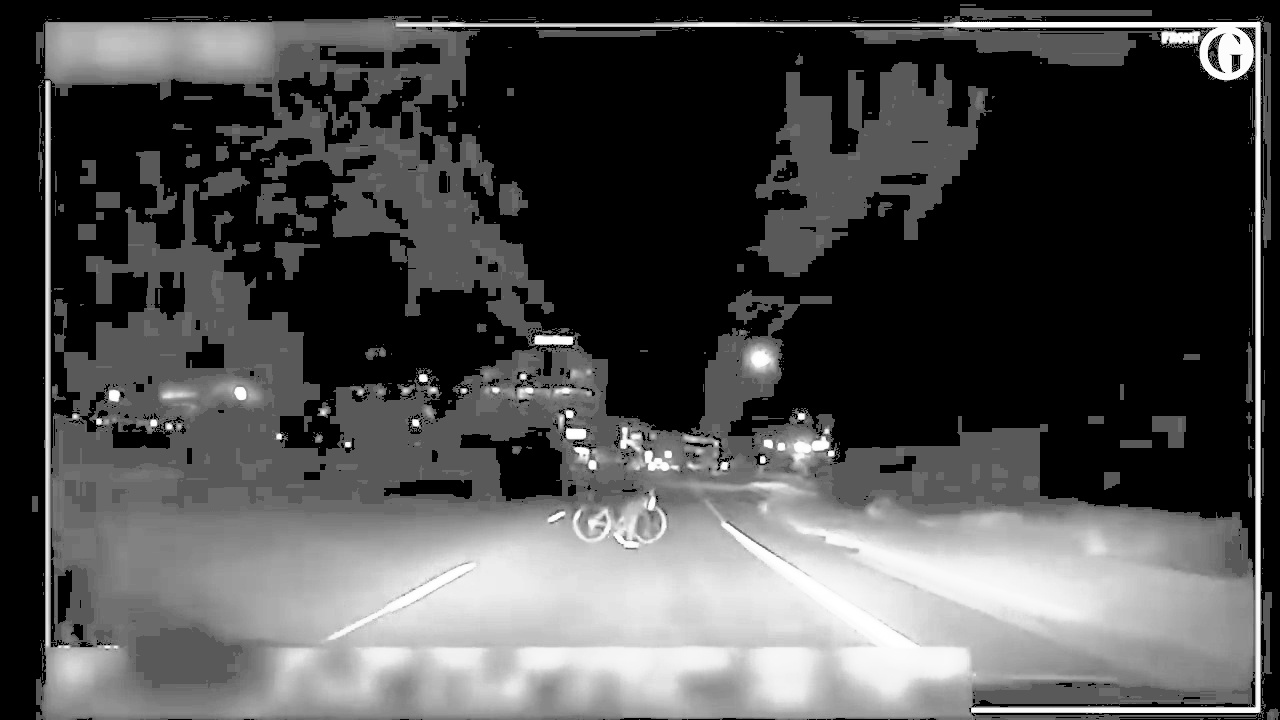}
\label{fig:testsfigure2}
\caption{Histogram Equalized image for Frame 75}
\end{figure}
   \subsubsection{Histogram Equalization} 
   
   While Gamma Correction modifies the luminance of an image, Histogram Equalization plays around with the contrast of an image. Contrast is determined by the difference in the color and brightness of the object and other objects within same field of view.
   
   Histogram Equalization achieves contrast enhancement by equalizing the image intensities. Multiple works have successfully used this technique for vision enhancement \cite{HE2clahe, HE4clahe} and object detection \cite{HE1}, and medical image processing \cite{HE5clahe}. By redistributing the pixels between the highest and the darkest portions of an image, we make a dark image (underexposed) less dark and a bright image (overexposed) less bright.
   
   Since Histogram Equalization considers the global contrast of the image, it doesn't lead to better image for our scenario as there is large intensity variations in every frame of the video. In other words, histogram covers a large region, i.e. both bright and dark pixels are present. To overcome this problem, we used adaptive Histogram Equalization method - CLAHE, where we divide every image into small blocks. For each of these blocks, we perform histogram equalization.

\begin{figure}[t]
\centering
\includegraphics[width=1\linewidth, ]{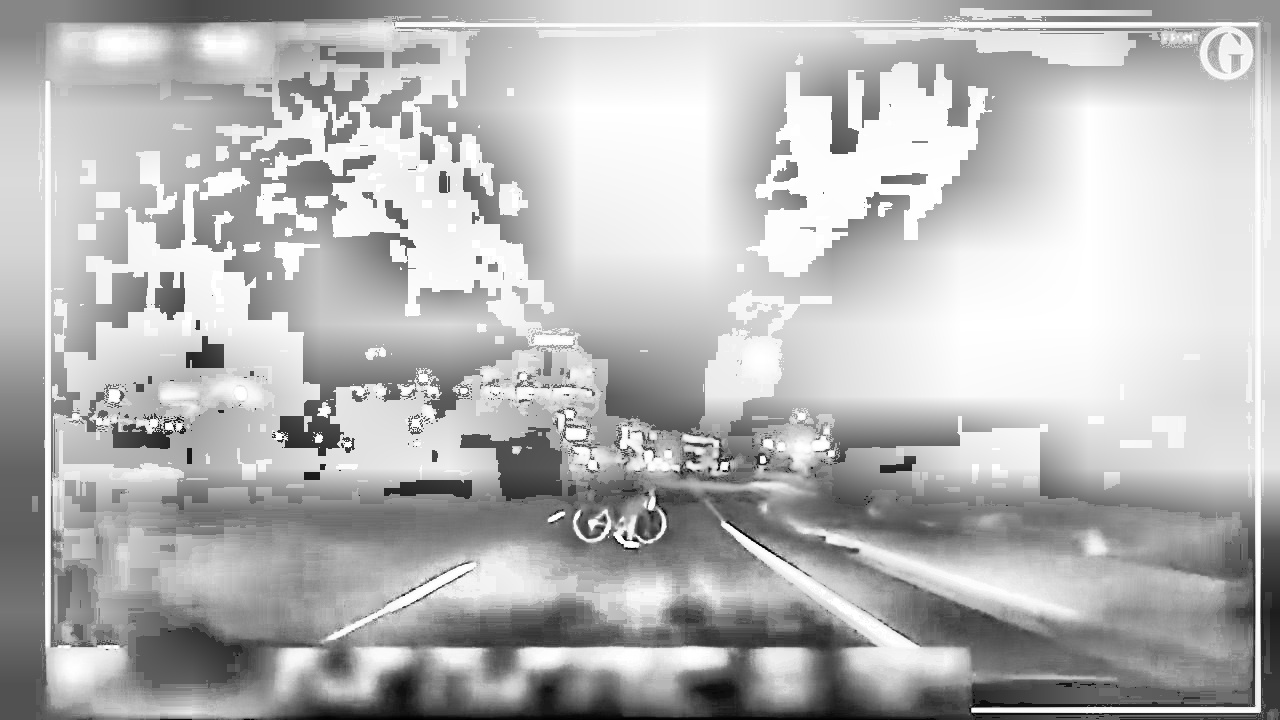}
\label{fig:testsfigure3}
\caption{Results after applying CLAHE to Frame 75}
\end{figure}

    \subsubsection{Edge Detection} 
    Multiple works have used Canny Edge Detection algorithm for pedestrian detection \cite{Edge5, Edge6, Edge7}. We applied same approach to create an edge map for every frame of video. A sample of such edge map is Figure 8.
    
    Canny Edge Detection is a multi-stage algorithm involving multiple steps --- image smoothing using Gaussian convolution, apply 2-D first derivative operator, perform non-maximal suppression. The effect of Canny operator is determined by --- width of the Gaussian kernel used while smoothing, the upper and lower thresholds used by tracker.

    \subsubsection{Adaptive Thresholding} 
    Tian \textit{et al.} \cite{AT1} introduced an adaptive thresholding segmentation algorithm for nighttime pedestrian detection.They follow a two-step approach - a detection phase which performs image segmentation followed by recognition phase to classify whether the object is a pedestrian using Support Vector Machines (SVM). We implemented a slightly modified version of their recognition algorithm in order to detect pedestrians. Our version is as follows:-
    \begin{enumerate}
        \setlength{\itemsep}{0pt}
        \item Convert the input frame to Grey-scale 
        \item Identify all the Connected Components in the frame using Block Based Component Labelling (BBDT) \cite{chang2015block}
        \item Remove Components that are greater than 10000 or less than 50 pixels in size
        \item Remove Components whose bottom-most pixels are in the top or bottom 10\% of the frame
        \item Remove Components whose area ratio (ratio of actual area to area of bounding box) is less than 0.5
    \end{enumerate}
    
    Figure 6 shows the comparison of our method against \cite{AT1} on a sample pedestrian image. Figure 7 shows the results for this method on Uber video frame 75.
    
    \begin{figure}[t]
    \centering
    \includegraphics[width=1\linewidth, ]{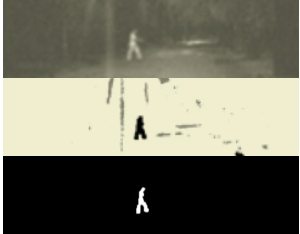}
    \label{fig:adaptive_compare_fig}
    \caption{Comparison of Adaptive Thresholding Segmentation algorithms. The image on the top is the original image, the one in the center is from \cite{tian2015pedestrian}, and the one on the bottom is using our algorithm. }
    \end{figure}  
    
    \begin{figure}[t]
    \centering
    \includegraphics[width=1\linewidth, ]{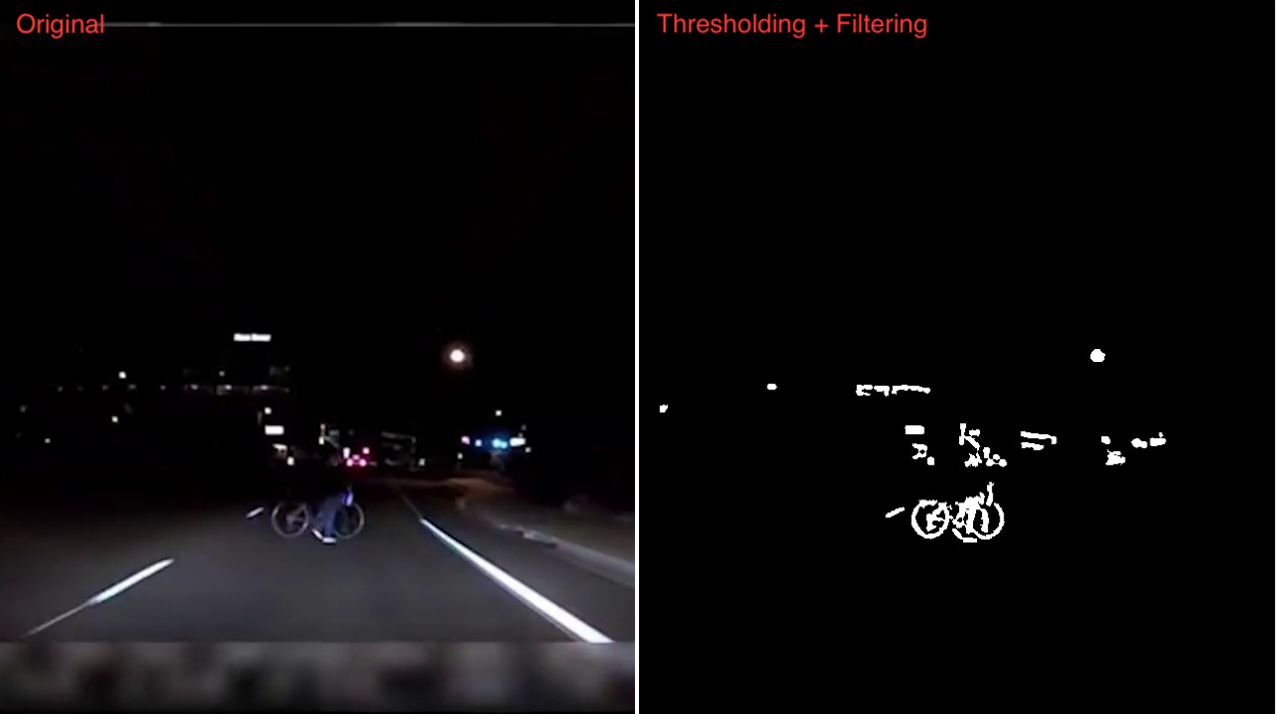}
    \label{fig:testsfigure4}
    \caption{Adaptive Thresholding Segmentation on Frame 75. The bicycle is outlined almost perfectly but the image is not devoid of noise. Our algorithm barely works on most other frames, completely failing to identify any object.}
    \end{figure}
    \subsubsection{Multi-Exposure Fusion Framework} 
    
    Ying et al. \cite{ying2017new} proposed a new Retinex based model which uses a dual-exposure fusion framework to enhance constrast and lightness of images. They create an enhanced image by fusing the original image with a synthetic image. This method has achieved state-of-the-art results, and we experimented with it for our image enhancement step.

\subsubsection{Camera Response Model} 

Ying et al. \cite{ying1} proposed a new image enhancement method which uses the response characteristics of cameras. We implemented this approach to check its effectiveness in improving the lighting condition of our video. First, we investigated the relationship between two images with different exposures to obtain an accurate camera response model. Following this, we borrowed the illumination estimation techniques to estimate the exposure ratio map. In the final step, this camera response model is used to adjust each pixel to its desired exposure according to the estimated exposure ratio map.
    
\subsection{Object Detection Approaches}
    \subsubsection{HOG + SVM} 
    
    This approach uses locally normalized HOG descriptor as features as described in the related work and uses linear svm as a baseline classifier. This approach provides good performance, in fact better than other existing feature sets including wavelets.
   
    \subsubsection{You Only Look Once (YOLO)} 
    
    YOLO \cite{yolov1, yoloV2, yolov3}  is one of the most promising state-of-the-art, real time object detection systems. What separates it from other object detector systems is the approach of applying a single neural network to the full image instead of applying the model to multiple locations of an image.
    Compared to other region proposal classification networks (fast RCNN) which perform detection on various region proposals and thus end up performing prediction multiple times for various regions in an image, YOLO architecture passes the image once through the`network and the output is a prediction.
    
    YOLO network divides the image into grid cells. Each of these grid cells is responsible for predicting 5 bounding boxes. A bounding box is described by a rectangle that encloses an object. YOLO also outputs a confidence score that informs us the certainity with which an object might be present in a grid cell. The YOLO architecture only uses standard layer types --- convolutional layers with a 3 * 3 kernel and max-pooling with a 2 * 2 kernel. In the second version of YOLO, the fully connected layers are removed. The very last convolutional layer has a 1 * 1 kernel.
    
    YOLO is written in Darknet. YOLO and other object detection models use Intersection over Union (IOU) as an evaluation metric. Any algorithm that provides predicted bounding boxes as output can be evaluated using IoU. This metric is simply a ratio of area of overlap between the predicted bounding box and the ground truth bounding box to the area of union between the predicted and ground truth boxes. 
    
    Multiple works have applied YOLO or model architectures similar to it for autonomous driving applications and have achieved good prediction accuracy \cite{yoloSqueeze, yoloRealTime, mvyolo}.
    
    \subsubsection{SSD} 
    
    Another popular deep learning based object detection method is Single Shot Detector (SSD) \cite{SSD}. Similar to YOLO, SSD uses a single network. However, it divides the image into a set of default boxes over different aspect ratios. For each default box, we predict the confidence for all object categories and adjusts the box location to better match the object location. For the task of pedestrian detection, SSD gives high accuracy on par with the state-of the-art object detectors \cite{SSD2}.
    
    \subsubsection{RetinaNet}
    
    Similar to SSD and YOLO, RetinaNet is a one-stage detector \cite{retinanet} which beats other two-stage detectors like Faster-RCNN \cite{LRfasterRcnn} in performance for object detection. The paper introduces Focal Loss --- a new loss function for classification, which increases the performance significantly. This loss function replaces the cross entropy loss function and is targeted to solve the problem of extreme foreground-background class imbalance which is encountered during training.

\begin{figure}[t]
\centering
\includegraphics[width=1\linewidth, ]{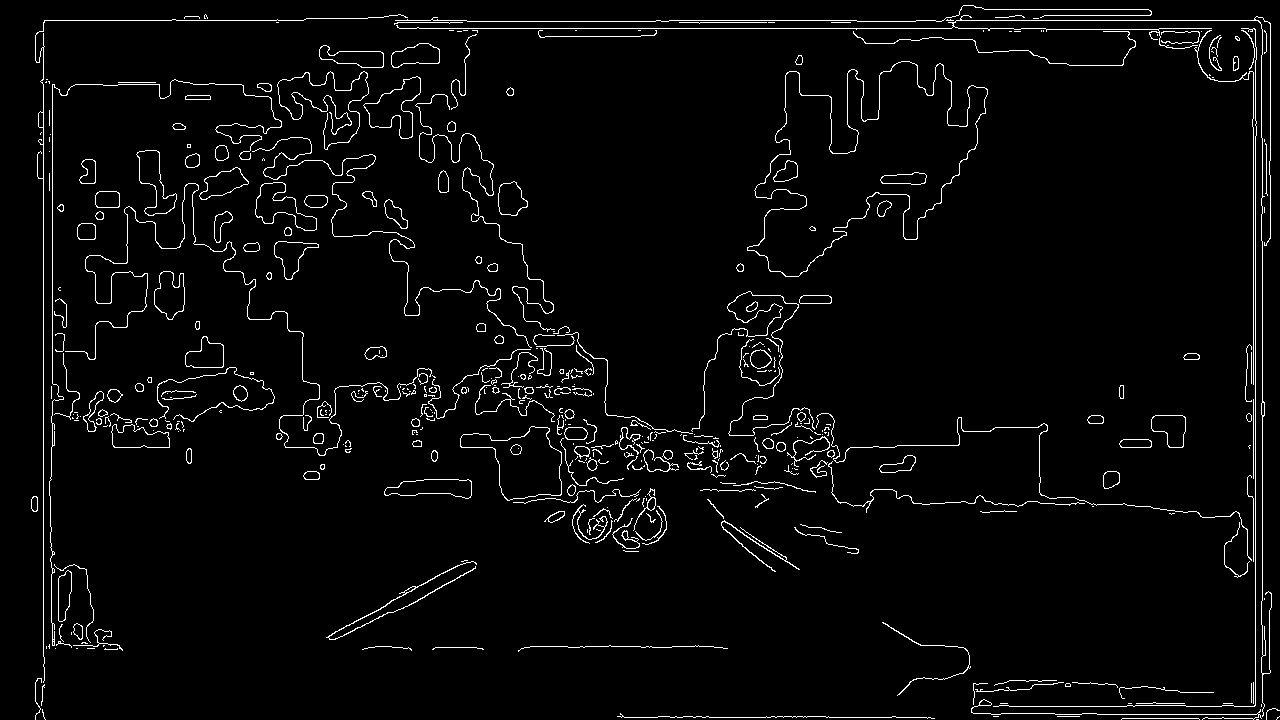}
\label{fig:canny_frame_75}
\caption{Canny Edge Detection on Frame 75}
\end{figure}  
\section{Experimentation}

\subsection {Experimental Settings}
    We ran all our experiments on Linux CentOS 7 machines with Intel Xeon E5-2680 v4 2.40GHz 14-core processors (128GB), and two NVIDIA Tesla K80 Graphical Processing Units (GPU). 

    The footage provided by Uber is shot at 24 frames per second. In order to detect the pedestrian in the Uber crash video, we converted the video into a total set of 165 frames, out of which every frame after frame 95 is the same. Frame 95 is where the actual crash happened. The pedestrian's feet become first visible at frame 60, and the complete silhouette is visible by frame 73. In other words, the crash happened at 3.95 seconds -- 0.91 seconds after the pedestrian was first fully visible, and 1.45 seconds after the pedestrian's feet were visible. As per a subjective analysis, we deduce that MobileEye was able to identify the pedestrian at frame 74 i.e. 0.86 seconds before the crash.

\subsection{Results}
Table \ref{tab:processing_time} shows the time taken to process the complete 165 frames of the input video, as well as the FPS for each of the Image Processing and Enhancement technique we applied to the video. As is visible, the more "basic" techniques such as Histogram Equalization, Canny Edge Detection, Gamma Correction, and Binary Thresholding were able to process the whole video at over 100 frames per second. On the other hand, more robust models such as the Camera Response Model and the Multi-Exposure Fusion Framework take upto 1 second per frame. This already gives us some intuition that for a real-time system that could potentially be deployed in an autonomous vehicle, we would need to stick to simpler image processing techniques that can process frames extremely fast.
\def\tabularxcolumn#1{m{#1}}
\begin{table}[tp]
\begin{center}
\begin{tabularx}{\linewidth}{|*1{>{\centering\arraybackslash}X}| *2{>{\centering\arraybackslash}c|}}
\hline

\textbf{\textit{Method}} & \textbf{\textit{Time (sec)}}& \textbf{\textit{FPS}} \\
\hline
Histogram Equalization          & 0.85    & 192.41  \\\hline
Canny Edge Detection                      & 1.00    & 165.16  \\\hline
Binary Thresholding             & 1.14    & 145.16  \\\hline
Gamma Correction                & 1.51    & 115.98  \\\hline
CLAHE                           & 1.72    & 96.00   \\\hline
Adaptive Threshold Segmentation    & 5.08    & 32.51   \\\hline
Motion Map (Shadows)    & 9.90    & 16.56  \\\hline
Motion Map (No Shadows) & 10.42   & 15.75  \\\hline
Harris Corner Detection                 & 50.50   & 3.27  \\\hline
Camera Response Model           & 136.37  & 1.21 \\\hline
Multi-Exposure Fusion Framework       & 164.55  & 1.00    \\\hline
\end{tabularx}
\label{tab:processing_time}
\end{center}
\caption{Processing Time of Image Enhancement Techniques.}
\end{table}

Table \ref{tab:detect_results} shows the amount of time taken for the various Object Recognition models to process the video. We also report which frame the person was first detected in. Both RetinaNet and YOLO were able to detect the pedestrian at frame 74, but RetinaNet's framerate was astonishingly the worst among our chosen models, processing only 1.57 frames per seconds. A surprising result was that the classical HOG + SVM based approach was able to successfully detect the pedestrian at frame 75, only one frame after YOLO and RetinaNet. The drawback of this approach was that there were too many false positives throughout the video. SSD nearly failed to recognize the pedestrian, just barely making it at frame 90, a few milliseconds before the actual crash. As the Uber video is a challenging video due to the unknown downsampling, low quality, and low-light scenario, we also tested our models against numerious other videos, and also against specific frames from the Caltech Pedestrian Dataset. On this dataset, we observed a similar pattern in the accuracies and processing time for each model.

It is clear that out of the chosen models, YOLO outperforms the others both in terms of accuracy of detection as well as processing time. By running YOLO on the raw input video frames, we were able to detect the pedestrian at frame 74, on par with the proprietary MobileEye model by Intel. In terms of time loss, this is \textbf{0.86 seconds before the actual crash}.

As YOLO was our best performing model, we ran the enhanced set of frames only on this model. All of the techniques described in Section 3.1 had no improvement on the detection of the pedestrian, with the best detection capped at frame 74. Interestingly, most of the enhancement techniques reduced the detection accuracy, with YOLO often failing to detect the pedestrian intermittently on the enhanced image sets.

\def\tabularxcolumn#1{m{#1}}
\begin{table}[tp]
\begin{center}
\begin{tabularx}{\linewidth}{|c | *3{>{\centering\arraybackslash}X|}@{}}
\hline
\textbf{\textit{Method}} & \textbf{\textit{Person Detected (Frame \#)}}& \textbf{\textit{Processing Time (sec)}} &\textbf{\textit{FPS}}\\
\hline
\textbf{RetinaNet}   & 74 &  105.46 & 1.57\\\hline
\textbf{HOG + SVM}   & 75 &  44.23  & 3.73\\\hline
\textbf{SSD}         & 90 &  33.36  & 4.94\\\hline
\textbf{YOLO}        & \textbf{74} &  \textbf{18.59}  & \textbf{8.87}\\\hline
\end{tabularx}
\label{tab:detect_results}
\end{center}
\caption{Results. YOLO outperforms all other models in terms of both accuracy and speed of detection. SSD Performs the words, whereas RetinaNet has comparable accuracy but very slow speed.}
\end{table}

%

\section{Conclusion}

In this work we showed that the YOLO Object Detection framework is able to detect the pedestrian from the Uber crash approximately one second before the actual crash. Our results with YOLO are on par with the proprietary models that have claimed similar results. Though we tried a variety of image enhancement and processing techniques, neither of them could further improve the detection rate on the Uber video. Our experiments on other videos and datasets were limited, but we are confident that in a general low-lighting scenario, these techniques would be beneficial for enabling pedestrian safety. 

While we still do not have a concrete answer to the original question -- "Could the Uber Car Crash have been avoided?", through our experiments we have shown that through a variety of available object detection models, we could successfully detect the pedestrian much before the actual accident. Whether the crash could have actually been avoided is a question best left for Uber to answer, as other factors such as decision making time, emergency braking system performance, etc. would also come into play. It is only a matter of time before Uber reveals what went wrong in their object recognition system and why the pedestrian could not be detected.

\begin{figure}[t]
\centering
\includegraphics[width=1\linewidth, ]{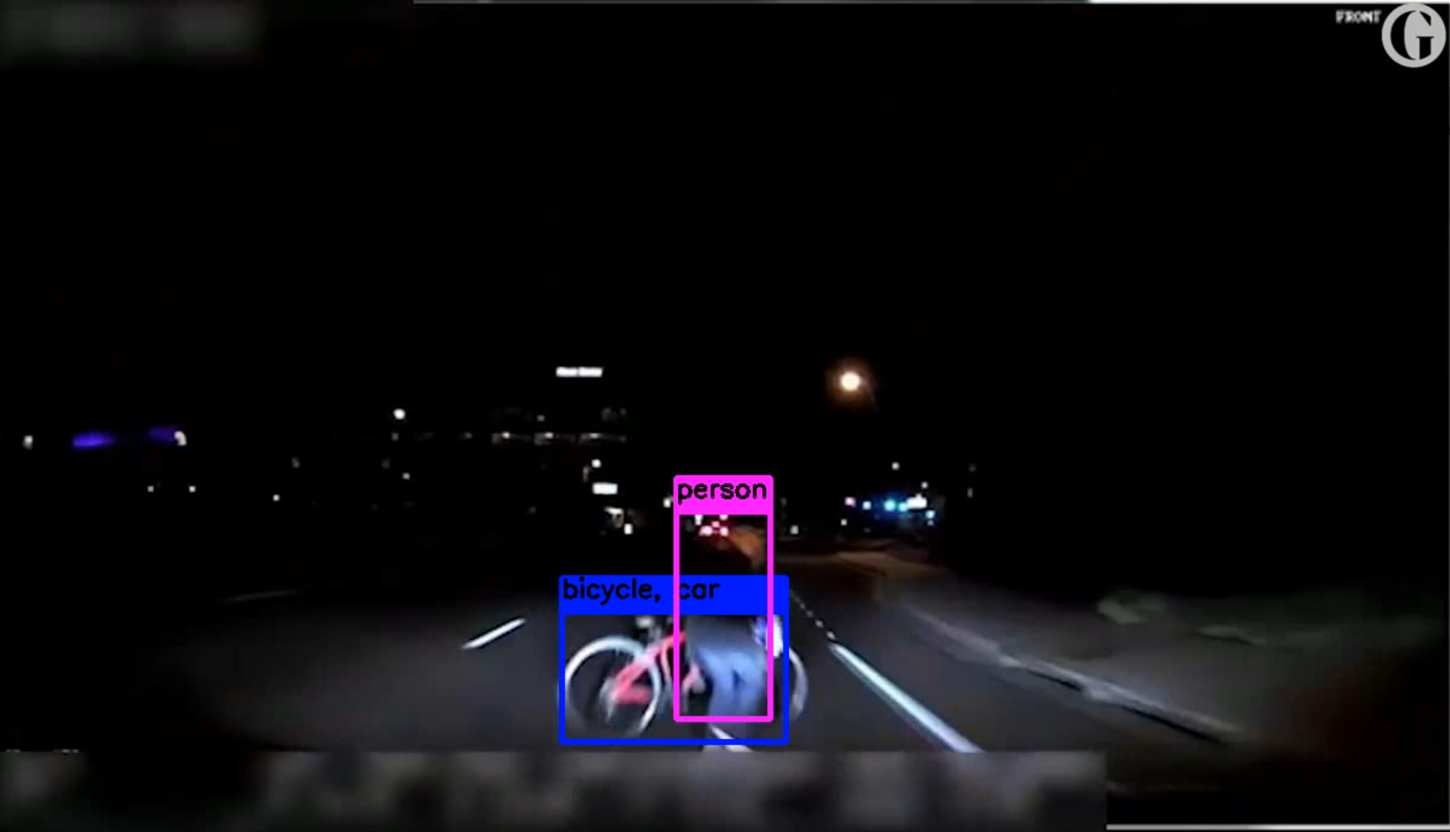}
\label{fig:testsfigure5}
\caption{YOLO detection on frame 87. YOLO was able to detect the pedestrian as well as the bicycle.}
\end{figure}
\section{Future Work}

Currently our image enhancement step is not directly linked to the object recognition models in a single pipeline. We first perform the image enhancement step and manually send the output to the recognition frameworks. As an immediate goal, we would like to create a joint model that processes low-light video frames and performs object recognition in real time. We aim to supplement this pipeline with a 'day/night' classifier that would also be able to judge whether or not low-light enhancement is required on an input video stream. Investigation of various permutations and sequencing of image enhancement techniques would also be a potential future work, aiming to visually enhance the pedestrians in a scene while filtering out unwanted background content.\newline
Further work would include fine-tuning our model specifically for pedestrian detection under various low-lighting scenarios. This would involve creating a synthetic dataset for pedestrian detection in low light conditions, done via augmenting the Caltech Pedestrian Dataset. We expect this to have a significant improvement in accuracy of detection in low lighting conditions. 


{\small
\bibliographystyle{unsrt}
\bibliography{uber_paper}
}

\end{document}